\documentclass{article}

\usepackage{microtype}
\usepackage{graphicx}
\usepackage{booktabs} 
\usepackage[caption=false]{subfig}

\usepackage{hyperref}

 \usepackage[accepted]{synsml2023}

\usepackage{amsmath}
\usepackage{amssymb}
\usepackage{mathtools}
\usepackage{amsthm}

\usepackage[capitalize,noabbrev]{cleveref}

\theoremstyle{plain}

\theoremstyle{definition}

\theoremstyle{remark}

\usepackage[textsize=tiny]{todonotes}

\synsmltitlerunning{Physics-Constrained Random Forests for Turbulence Model Uncertainty Estimation}

\begin{document}

\twocolumn[
\synsmltitle{Physics-Constrained Random Forests \\
            For Turbulence Model Uncertainty Estimation}



\synsmlsetsymbol{equal}{*}

\begin{synsmlauthorlist}
\synsmlauthor{Matha Marcel}{equal,yyy}
\synsmlauthor{Christian Morsbach}{yyy}
\end{synsmlauthorlist}

\synsmlaffiliation{yyy}{German Aerospace Center (DLR), Linder Höhe, 51147 Cologne, Germany}

\synsmlcorrespondingauthor{Marcel Matha}{marcel.matha@dlr.de}

\synsmlkeywords{Machine Learning, turbulence modeling, RANS, CFD, Random Forest, Uncertainty Quantification}

\vskip 0.3in
]



\printAffiliationsAndNotice{}  

\begin{abstract}
To achieve virtual certification for industrial design, quantifying the uncertainties in simulation-driven processes is crucial. We discuss a physics-constrained approach to account for epistemic uncertainty of turbulence models. In order to eliminate user input, we incorporate a data-driven machine learning strategy. In addition to it, our study focuses on developing an a priori estimation of prediction confidence when accurate data is scarce.
\end{abstract}

\section{Introduction}
\label{introduction}

In the past, the interest in Uncertainty Quantification (UQ) for Computational Fluid Dynamics (CFD) simulations has increased, leading to more reliable simulation-based engineering designs \cite{Karniadakis, Oberkampf, XIAO20191}. Reynolds-averaged Navier-Stokes (RANS) turbulence modeling is still widely used in industrial design due to its computational efficiency. However, RANS equations have an unresolved term called the Reynolds stress tensor, requiring approximation using turbulence models. The accuracy of these models is crucial but often limited \cite{Speziale, Mompean, CRAFT1996108}. Simplifications in turbulence model formulation introduce significant epistemic uncertainties.\\
The Eigenspace Perturbation Framework (EPF) tries to estimate the model-form uncertainty of the the turbulence model and helps optimize design with reduced sensitivity to uncertainty \cite{Emory, Iaccarino, Mishra2019}. It has been successfully applied in aerospace, turbomachinery, civil structural, and wind farm design \cite{mishra2017uncertainty, Cook, Mishra2020Design, mishra2019uncertainty, Lamberti, EidiDataFree}. Data-driven approaches, aided by high-fidelity simulations and Machine Learning (ML), are gaining popularity in RANS turbulence modeling \cite{Heyse, matha2022applicability}.
The present study, which was the foundation of the CFD application results in our previous paper \cite{MathaCF}, focuses on the use of data-driven methods in order to identify flow regions with potential turbulence model prediction inaccuracies. Random forests (RF) are trained and validated using additional data from scale-resolving simulations. We analyze a methodology to assess a priori prediction confidence. Our framework is an amalgam of physics based domain knowledge with data-driven RF.

\section{Eigenspace Perturbation Framework}
\label{sec_eigenspaceFramework}
The Eigenspace Perturbation approach aims to provide understandable uncertainty bounds for modeling the Reynolds stress tensor based on physical principles.
The Reynolds stress tensor can be decomposed as

\begin{equation}
	\label{spectralDecompositionR}
		\tau_{ij} = k \left(a_{ij} + \frac{2}{3}\delta_{ij}\right) = k \left(v_{in} \Lambda_{nl} v_{jl} + \frac{2}{3}\delta_{ij}\right) \ \text{,}
\end{equation} 
with $k = \frac{1}{2}\tau_{ii}$ being the turbulent kinetic energy.
The eigenspace decomposition provides the eigenvector matrix $v$ and the diagonal eigenvalue matrix $\Lambda$.
Emory et al. \yrcite{Emory} propose a strategy to perturb the eigenvalues in \Cref{spectralDecompositionR}, resulting in a perturbed state of the Reynolds stresses

\begin{equation}
	\label{spectralDecompositionR*}
		\tau_{ij}^* = k \left(v_{in} \Lambda_{nl}^* v_{jl} + \frac{2}{3}\delta_{ij}\right) \ \text{.}
\end{equation}

\begin{figure}
\centering
\begin{tikzpicture}
\draw [-](-2,-1.5) -- (2,-1.5) node[] {};
\draw [-](-2,-1.5) -- (0,1.964) node[] {};
\draw [-]( 2,-1.5) -- (0,1.964) node[] {};
\draw [dashed](-0.3, 0.1) circle (1.3cm);
\draw [gray, dashed]( -0.3, 0.1) -- (0,1.964) node[] {};
\draw [gray, dashed]( -0.3, 0.1) -- (-2,-1.5) node[] {};
\draw [gray, dashed]( -0.3, 0.1) -- (2,-1.5) node[] {};
\draw [blue, line width=0.5mm]( -1.24, -0.8) -- (-0.3, 0.1) node[] {};
\draw [red, line width=0.5mm]( 0.22, -1.1) -- (-0.3, 0.1) node[] {};
\draw [black, line width=0.3mm](0.76, -0.66) -- (-0.3, 0.1) node[] {};
\draw [black, line width=0.3mm](-0.09, 1.36) -- (-0.3, 0.1) node[] {};
\filldraw[black] (0.76, -0.66) circle (1pt) node[]{};
\filldraw[black] (-0.09, 1.36) circle (1pt) node[]{};
\filldraw[black] (1.7, -1.5) circle (0.01pt) node[anchor=north west]{$\mathbf{x}_{\mathrm{1C}}$};
\filldraw[black] (-1.7, -1.5) circle (0.01pt) node[anchor=north east]{$\mathbf{x}_{\mathrm{2C}}$};
\filldraw[black] (0, 1.964) circle (0.01pt) node[anchor=south]{$\mathbf{x}_{\mathrm{3C}}$};
\filldraw[black] (-0.3, 0.1) circle (2pt) node[anchor=west]{$\mathbf{x}_{\mathrm{RANS}}$};
\filldraw[black] (  -1.24, -0.8) circle (2pt) node[anchor=west]{$\mathbf{x}^*_{\mathrm{RANS}}$};
\node[blue] at (-0.9, -0.05) {$\Delta_B$};
\filldraw[black] ( 0.22, -1.1) circle (2pt) node[anchor=west] {$\mathbf{x}_{\mathrm{Data}}$};
\node[red] at (-0.3, -0.45) {$p$};
\end{tikzpicture}
\caption{Barycentric triangle: schematic representation of the eigenvalue perturbation and definition of perturbation strength.}
\label{fig:baryCentricTriangle}
\end{figure}
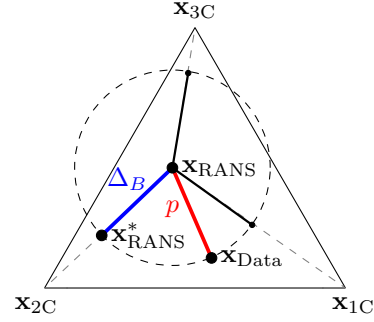

The eigenvalue perturbation, which involves determining $\Lambda^*$, builds on the concept that every physically realizable state of the Reynolds stress tensor can be represented by barycentric coordinates. In this representation, all realizable states of the Reynolds stress tensor lie within or on the boundaries of the barycentric triangle. The vertices of this triangle represent limiting states of turbulence (one- ($\mathrm{1C}$), two- ($\mathrm{2C}$) and three-component isotropic ($\mathrm{3C}$) state). A linear mapping is defined between these vertices and the eigenvalues $\lambda_1\geq\lambda_2 \geq\lambda_3$ of the Reynolds stress anisotropy \cite{Banerjee2007} (see \cref{fig:baryCentricTriangle}):
\begin{equation}
\label{barycentricMapping}
	\mathbf{x} = \mathbf{x}_{1C}\frac{\lambda_1-\lambda_2}{2}+ \mathbf{x}_{2}\left(\lambda_2-\lambda_3\right)+ \mathbf{x}_{3C} \left(\frac{3\lambda_3}{2}  +1\right) 
\end{equation}
The eigenvalue perturbation is defined as a shift in barycentric coordinates towards each of the limiting states $\mathbf{x}_{(t)} \in \{\mathbf{x}_{1C}, \mathbf{x}_{2C}, \mathbf{x}_{3C}\}$ by the relative distance $\Delta_B \in [0, 1]$:
\begin{equation}
	\label{perturbationMagnitude}
		\mathbf{x}^*_{\mathrm{RANS}} = \mathbf{x}_{\mathrm{RANS}} + \Delta_B \left(\mathbf{x}_{(t)} -\mathbf{x}_{\mathrm{RANS}}\right) \ \text{.}
\end{equation}
The final perturbed eigenvalues $\lambda_{i}^*$ can be remapped by inverting \cref{barycentricMapping} based on $\mathbf{x}^*_{\mathrm{RANS}}$.

Nevertheless, the uniform perturbation in the data-free procedure fails to capture the varying discrepancies between CFD's turbulence model based results and accurate turbulence physics across different flows and regions. Variation in the perturbation magnitude reflects the true model-form uncertainty more approriate, resulting in precise and less conservative uncertainty bounds for Quantities of Interest (QoI) \cite{EmoryThesis, Heyse, MathaCF}.
Therefore, we introduce a local perturbation strengths
\begin{equation}
    \label{eq:perturbationMagnitudeP}
    p = |\mathbf{x}_{\mathrm{Data}} - \mathbf{x}_{\mathrm{RANS}}| = |\mathbf{x}^*_{\mathrm{RANS}} - \mathbf{x}_{\mathrm{RANS}}| \ \text{,}
\end{equation}
as illustrated in \Cref{fig:baryCentricTriangle}.
The perturbation strength $p$ should be predicted by a ML model in order to modify the Reynolds stress towards the same three limiting states as in the physics-constrained data-free approach.
Hereby, on top of the foundation of the EPF, which is allowing for understandable uncertainty estimates for turbulence modeling, we incorporate data-driven ML techniques to enhance the predictive capabilities of the entire framework.
The final application of propagating these perturbations throughout the RANS solver is beyond the scope of this paper (for details on that see \cite{Heyse, MathaCF}).
Additionally, ML predicted $p$, representing the deviation of barycentric coordinates from high-fidelity data to RANS simulations, might be helpful for future turbulence modeling activities aiming for correct representation of Reynolds stress anisotropy \cite{Duraisamy1}.

\section{Application of Random Forests}
In this work, the python library \textit{scikit-learn} \cite{scikit-learn} is used to train RF and evaluate their predicted perturbation strength $p$.
\subsection{Choice of flow features}
The selection of input features is crucial for accurately predictions. It is important to choose features that have a meaningful physical significance and are relevant to the desired target quantity. Therefore, we agree on using four input quantities $Q = \left(\mathbf{S}, \mathbf{\Omega}, \nabla p, \nabla k\right)$, whereby the input tensors $\mathbf{S}$, $\mathbf{\Omega}$ represent strain rate and rotation rate, while $\nabla p$ and $\nabla k$ are the gradients of pressure and turbulent kinetic energy  \cite{Wang}.
Normalizing the raw input quantities by a factor $\beta$ and the absolute value of each element $\alpha$ according to $\hat{\alpha} = \frac{\alpha}{|\alpha|+|\beta|}$ \cite{Ling}, lead to the determination of non-dimensional flow features, which are presented in \Cref{tab:rawFlowFeatures}.

\begin{table}[h]
  \caption{Raw flow features for constructing the invariant basis}
  \label{tab:rawFlowFeatures}
  \centering
  {\footnotesize
  \begin{tabular}{c c c}
    \toprule
    Normalized input $\hat{\alpha}$& raw input $\alpha$ & normalization factor $\beta$  \\
    \midrule
    $\mathbf{\hat{S}}$      &$\mathbf{S}$         &  $\omega$\\
    $\mathbf{\hat{\Omega}}$ &$\mathbf{\Omega}$    & $||\mathbf{\Omega}||$\\
    $\widehat{\nabla p}$        &$\nabla p$           & $\rho ||\mathbf{U} \cdot \nabla \mathbf{U}||$\\
    $\widehat{\nabla k}$        &$\nabla k$          & $\omega \sqrt{k}$\\
    \bottomrule
  \end{tabular}
  }
\end{table}

\begin{table}[b]
  \caption{Physical flow features}
  \label{tab:physicalFeature}
  \centering
  {\footnotesize
  \begin{tabular}{c c c c}
    \toprule
    Numbering & raw input $\alpha$ & normalization factor $\beta$  \\[1ex]
    \midrule
    $q_1$   &$\frac{1}{2}\left(||\mathbf{\Omega}||^2 - ||\mathbf{S}||^2\right)$         &  $||\mathbf{S}||^2$\\[1ex]
    $q_2$ & $k$& $\frac{1}{2}U_i U_i$\\[1ex]
    $q_3$        &$\min\left(\frac{\sqrt{k}d}{50\nu},2\right)$           & -\\[1ex]
    $q_4$       &$U_k\frac{\partial p}{\partial x_k}$          & $\sqrt{\frac{\partial p}{\partial x_j}\frac{\partial p}{\partial x_j}U_iU_i}$\\[1ex]
    $q_5$       &$\frac{1}{\omega}$          & $\frac{1}{||\mathbf{S}||}$\\[1ex]
    $q_6$       &$P_k$          & $k\omega$\\[1ex]
    $q_7$        & $Ma$          & -\\[1ex]
    $q_8$        & $\mu_t$          & $\mu$\\[1ex]
    $q_9$        & $||\overline{u_i'u_j'}||$        & k\\
    \bottomrule
  \end{tabular}
  }
\end{table}

\begin{table*}[htb]
  \caption{Accuracy of RF: $x$ means part of training data, $\circ$ means \underline{not} part of training data, \textcolor{red}{red} means data set for evaluation of $\mathrm{RMSE}$}
  \label{tab:predictionAccuracy}
  \hspace*{-1.2cm}
  \centering
  {\footnotesize
  \begin{tabular}{llllllllllll}
    \toprule
                       &  &       \multicolumn{10}{c}{Scenario}                   \\
    \cmidrule(r){3-12}
    \multicolumn{2}{l}{Flow cases}          & I     & Ia & II & IIa & III & IIIa & IIIb & IIII & IIIIa& V\\
    \midrule
    \multicolumn{2}{l}{turbulent channel}                 & x                        & x                   & x                          & x                  & x                         & x            &x     &\textcolor{red}{$\circ$} & \textcolor{red}{x}&\textcolor{red}{x}\\
    \multicolumn{2}{l}{periodic hill}                     &&&&&&&&&\\
    & \footnotesize{$\cdot \ Re_H \in \{2800, 5600, 10595\}$}  & x                        & x                   & x                          & x                  &\textcolor{red}{$\circ$}   &\textcolor{red}{x} & &x&x&\textcolor{red}{x}\\
    &\footnotesize{$\cdot \ Re_H \in \{2800, 10595\}$}  &   &  & &  &   & &x &&\\
    &\footnotesize{$\cdot \ Re_H = 5600$}  &   &  & &  &   & &\textcolor{red}{$\circ$} &&&\\
    \multicolumn{2}{l}{wavy wall}                         & x                        & x                   & \textcolor{red}{$\circ$}    & \textcolor{red}{x} &x                          &x&x &x&x&\textcolor{red}{x}\\
    \multicolumn{2}{l}{converging-diverging channel}      & \textcolor{red}{$\circ$} & \textcolor{red}{x}  & x                          & x                  &x                          &x&x &x&x&\textcolor{red}{x}\\
    \midrule
    \multicolumn{2}{l}{$\mathrm{RMSE}\left(p_{\mathrm{pred}},p_{\mathrm{true}}\right)$} & 0.098 & 0.010 & 0.133 & 0.029&  0.095&0.028& 0.041& 0.051& 0.014 & 0.013     \\
    \bottomrule
  \end{tabular}
  }
\end{table*}
Inspired by the work of Wang et al. \yrcite{Wang}, we determine the integrity feature basis based on the non-dimensional raw flow features resulting in 47 invariant features. These invariants, along with additional physically meaningful flow features, serve as input for training and evaluating the RF. The additional features presented in \Cref{tab:physicalFeature} include variables such as turbulent kinetic energy $k$, specific turbulent dissipation rate $\omega$, molecular viscosity $\mu$, eddy viscosity $\mu_t$, distance to the nearest wall $d$, local Mach number $Ma$, mean velocity $U_i$ and its gradient tensor, and mean pressure $p$ and its gradient vector.

A total number of 56 input features is used for training and evaluating the RF.
Finally, to standardize each feature, the mean is subtracted and the data is scaled to have unit variance.

\subsection{Data sets}
\label{sec_dataSets}
For the present study, we use the following high-fidelity simulations (Direct Numerical Simulation (DNS) and Large Eddy Simulations (LES)) as training and testing data sets:
\vspace{-0.2cm}
\begin{itemize}
\itemsep0em 
    \item DNS of turbulent channel flow at $Re_\tau \in \{180, 550, 1000, 2000, 5200\}$ \cite{lee_moser_2015}
    \item DNS at $Re_H \in \{2800, 5600\}$ and LES at $Re_H = 10595$ of periodic hill flow \cite{Breuer2009}
    \item DNS of wavy wall flow at $Re_H = 6850$ \cite{Rossi2006}
    \item DNS of converging-diverging channel flow at $Re_\tau=617$ \cite{Laval}.
\end{itemize}
\vspace{-0.2cm}
While the input features are extracted based on separately performed RANS simulations, the local perturbation magnitude $p$ is determined according to \Cref{eq:perturbationMagnitudeP} by comparing these RANS results and high-fidelity simulations. 

\subsection{Verification of ML model}
\label{sec_Modelvalidation}
\begin{figure*}[t]
\captionsetup[subfloat]{farskip=1pt,captionskip=0.5pt}
\centering
\hspace*{1.5cm}
\begin{subfloat}[\label{convDivMetric} Evaluated extrapolation metric taking into account the features $q_1$, $q_2$, $q_3$, $q_7$ and $q_8$.]{\includegraphics[width=0.82\textwidth]{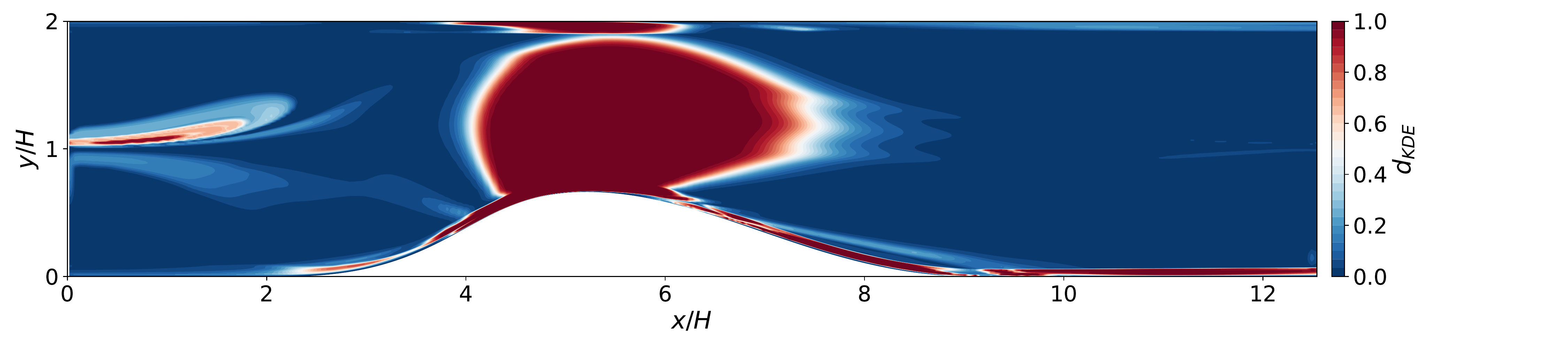}}
\end{subfloat}
\hspace*{1.7cm}
\begin{subfloat}[\label{convDivP} Model prediction error taking into account all 56 input features.]{\includegraphics[width=0.82\textwidth, trim=0cm 0.2cm 0cm 0cm, clip=True]{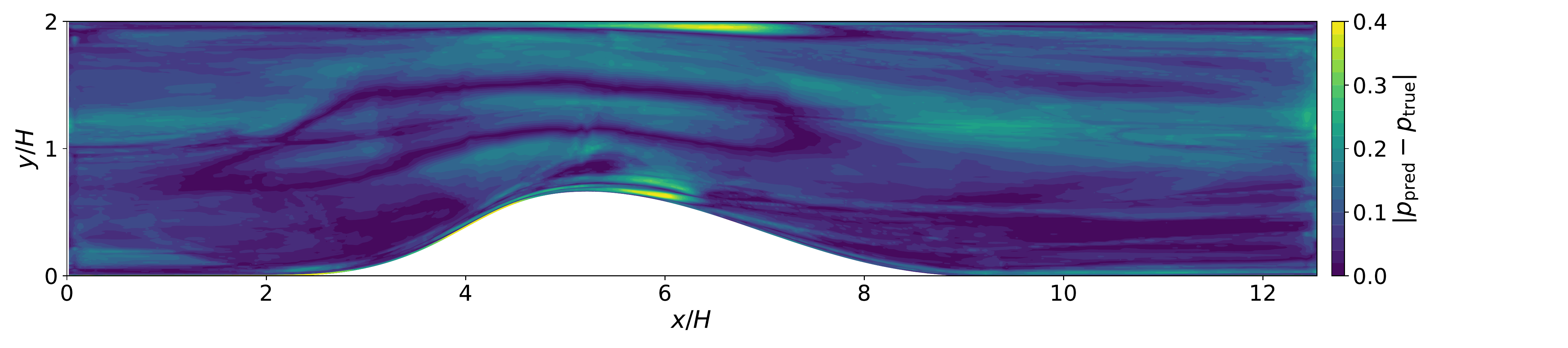}}
\end{subfloat}
\caption{Spatial comparison between KDE metric and RF error for flow of converging-diverging channel (scenario I in \Cref{tab:predictionAccuracy}}
\label{Bild:convDivMetric}
\end{figure*}
The prediction accuracy of the RF should be evaluated on the available data (see \Cref{sec_dataSets}) with fitted hyperparameters (\textit{maximum tree depth} = 15, \textit{minimum sample count} = 10, \textit{maximum number of features} = 7, \textit{number of trees} = 30) based on leave-one-out-cross validation (not shown here). Ten different scenarios ,combining individual flow cases listed in \Cref{tab:predictionAccuracy} for training and testing purposes, serve as verification of functionality and present the accomplishment of the intended generalization of every model. 
Since the target quantity $p$ ranges between zero and one (due to the construction of the equilateral triangle with an edge length of one), the resulting $\mathrm{RMSE}$ values indicate less than 10\% absolute prediction error, except for scenario II.

Evaluating the accuracy of a trained ML model becomes challenging when making predictions on flow cases without accurate data. To build confidence in the model's performance, it is reasonable to compare the input feature spaces of the training and testing data, which results in extrapolation metrics.
In this study, we apply the Kernel Density Estimation (KDE) to determine the extrapolation metric, measuring the distance between a test point $\tilde{m}$ and the training data feature set $m^{(i)}$ for $i = 1,\dots,n$ with $n$ as the number of training data points
\begin{align}
\label{eq:kdeDistance}
    d_{\mathrm{KDE}} &= 1-\frac{f_{\mathrm{KDE}}}{f_{\mathrm{KDE}}+1/A}
\end{align}
whereby $A=\prod_i^n \left(\max_j\left({m^{(i)}_{j}}\right) - \min_j\left({m^{(i)}_{j}}\right)\right)$ for $j = 1,\dots,d$.
The probability density is estimated via
\begin{equation}
\label{eq:kernel}
    f_{\mathrm{KDE}} = \frac{1}{n \sigma^{d}} \sum_{i=1}^n \prod_{j=1}^d K \left(\frac{\tilde{m}_j-m^{(i)}_{j}}{\sigma}\right) \ \text{,}
\end{equation}
with the number of features $d$, the bandwidth $\sigma$ (determined by Scott's rule \cite{Scott}) and a Gaussian kernel $K\left(t\right) = 1/\sqrt{\left(2 \pi \right)} \exp{\left(-t^2/2\right)}$.
The Gaussian kernel $K$ employed in the method ensures that as the difference between $\tilde{m}_j$ and $m^{(i)}_{j}$ decreases, the output of $K$ increases. In simpler terms, \Cref{eq:kernel} yields higher values when $\tilde{m}$ approaches a concentrated feature space of the training data points, and vice versa. 
By normalizing the distance in \cref{eq:kdeDistance}, the metric effectively measures the distance between $\tilde{m}$ and the training data with respect to a uniform distribution. Consequently, when $\tilde{m}$ is close to a concentrated feature space of $m^{(i)}_{j}$, $f_{KDE} \gg 1/A$ resulting in $d_{\mathrm{KDE}}$ approaching 0 (no extrapolation is required). Conversely, when $f_{KDE}$ is much smaller than $1/A$, $d_{\mathrm{KDE}}$ tends to approach 1 (high extrapolation is required).
This provides users with the ability to assess the degree of extrapolation required based on the characteristics of the training data set.

\begin{figure}[h]
\begin{center}
\includegraphics[scale=0.27, trim=0.5cm 0cm 0.7cm 0.1cm, clip=False]{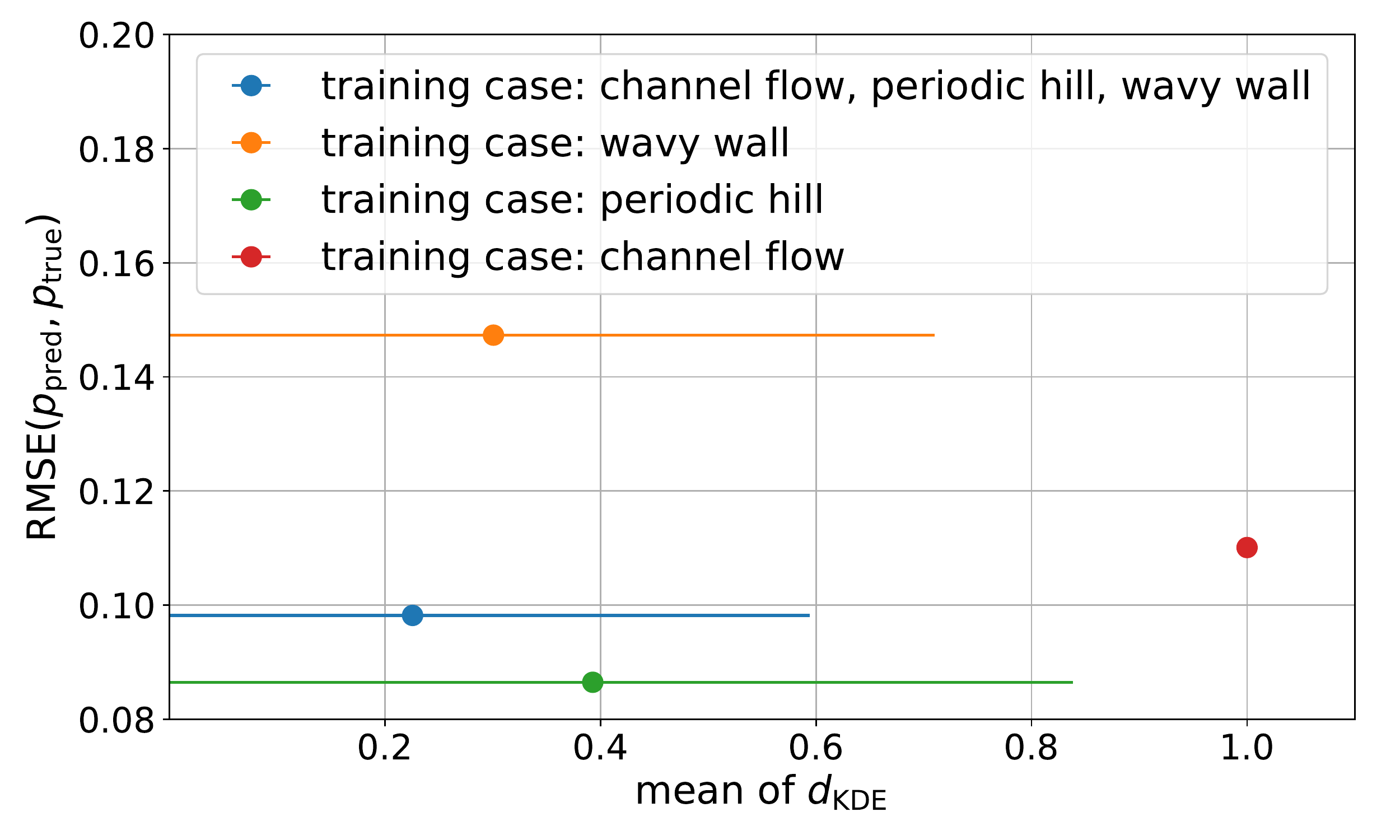}
\vspace{-1\baselineskip}
\caption{Relationship between the RMSE of the prediction and the mean of KDE metric for the converging-diverging channel (standard deviation as horizontal bars). All features are considered for the prediction and training of the RF models, while only $q_1$, $q_2$, $q_3$, $q_7$ and $q_8$ are used to compute $d_{\mathrm{KDE}}$.}
\label{Bild:convDivCorrelation}
\end{center}
\end{figure}

In our study, the converging-diverging channel flow case demonstrates the use of the extrapolation metric, while separate RF are trained on the different data sets. Key features with significant importance for the training are considered for calculating the KDE distance, determined using permutation feature importance. Though, in the definition of the KDE metric, all features are initially treated equally. The selected five most important features are $q_8$, $q_3$, $q_7$, $q_2$ and $q_1$. Unlike previous work \cite{Wu}, we do not observe a strong correlation between model accuracy in predicting perturbation magnitude for the converging-diverging channel and the mean KDE distance for different training data sets. (see \Cref{Bild:convDivCorrelation}).

The barycentric coordinates of the converging-diverging channel and periodic hill in the DNS data cover similar areas in the barycentric triangle. In contrast, the barycentric coordinates for the wavy wall test case are predominantly located in the lower range of the triangle. 
Consequently, the distance in barycentric coordinates, as the target quantity, tends to have more frequent values within a similar range for the converging-diverging channel and periodic hill. This difference could explain the reduced prediction error when training on the periodic hill compared to the wavy wall. Wu et al. \yrcite{Wu} also stated that the correlation between accuracy and the extrapolation metric is weaker when the training set is very similar or very different from the test set, which might be the case here as well.

\Cref{convDivMetric} presents the two-dimensional distribution of the KDE metric, evaluated based on features $q_8, q_3, q_7, q_2$ and $q_1$.
Although, some spatial correlated regions between $d_{\mathrm{KDE}}$ and the model errors in \Cref{convDivP} can be recognized, their overall correlation is not strong (Pearson correlation coefficient $\approx$ 0.2). Unfortunately, areas characterized by lower KDE metrics also exhibit less precise model predictions (e.g. $y/H\approx 1$ and $x/H>8$).

However, it is important to note that the outcome of the extrapolation metric heavily relies on the chosen set of features as all features are equally weighted. Therefore, it is reasonable to evaluate the metric only on a subset of important features for the RF.
\\[-2.9ex]
\section{Conclusion and Outlook}
\label{sec_concluison}
Our approach represents a hybrid framework that combines a physics-based UQ methodology with data-driven ML. Using ML predicted, spatially varying perturbation strength helps CFD practitioners to identify regions, in which the turbulence model might be erroneous on the one hand. Propagating these prediction to the RANS solver helps to obtain data-driven, less conservative and nonetheless physics-constrained uncertainty estimates for QoI on the other hand\cite{MathaCF}.\\
Additionally, we focused on important aspects in the field of ML, such as the selection of features and evaluating the accuracy of the model using a posteriori and a priori approaches. In particular, the a priori estimation of ML confidence based on the KDE extrapolation metric lacks correlation with the prediction error globally and locally. Nevertheless, we strongly believe that the evaluation of ML credibility holds significant importance for forthcoming CFD design optimization circles. Hence, it is imperative that future research related to ML application for CFD design optimization places emphasis on addressing this issue.



\section*{Broader impact}
Numerical analysis based on flow simulations using software has become increasingly important in industrial aerodynamic designs. These applications commonly involve turbulent flows.
Despite the significant increase in computational power over the years, achieving scale-resolving (high fidelity) simulations for design optimization studies in complex engineering designs remains a challenging task. The accuracy of cost-effective RANS simulations, which heavily rely on turbulence models, will continue to be the state-of-the-art approach in the coming years. The primary objective of this study is to consolidate the emerging methods in turbulence model uncertainty quantification and combine them with ML.
An appropriate application of ML to enhance the methodology to quantify turbulence model uncertainties will pave the way towards reliability based design or virtual certification. 




\nocite{langley00}

\bibliography{main}
\bibliographystyle{synsml2023}



\end{document}